%% file: main.tex
\documentclass[10pt,twocolumn,letterpaper]{article}

\usepackage{wacv}      % To produce the REVIEW version for the algorithms track
\input{macros}

\input{preamble}

\definecolor{wacvblue}{rgb}{0.21,0.49,0.74}
\usepackage[pagebackref,breaklinks,colorlinks,allcolors=wacvblue]{hyperref}

\def\wacvPaperID{2136} % *** Enter the WACV Paper ID here
\def\confName{WACV}
\def\confYear{2026}

\title{Boosting Unsupervised Video Instance Segmentation with \\ Automatic Quality-Guided Self-Training}

\author{
Kaixuan Lu$^{1}$ \quad Mehmet Onurcan Kaya$^{1,2}$ Dim~P.~Papadopoulos$^{1,2}$\\%[0.5em]
$^{1}$Technical University of Denmark \quad 
$^{2}$Pioneer Centre for AI\\%[0.5em]
\texttt{\small s232248@student.dtu.dk, monka@dtu.dk, dimp@dtu.dk}
}

\begin{document}
\maketitle
\input{sec/0_abstract}    
\input{sec/1_intro}
\input{sec/2_relatedwork}
\input{sec/3_methodology}
\input{sec/4_experiment}
\input{sec/5_conclusion}
\section*{Acknowledgements}
D. P. Papadopoulos and M. O. Kaya were supported by the DFF Sapere Aude Starting Grant
``ACHILLES''.
{
    \small
    \bibliographystyle{ieeenat_fullname}
    \bibliography{main}
}

\input{sec/X_suppl}

\end{document}

%% file: macros.tex
\iffalse
\newcommand{\dimpp}[1]{\textcolor{blue}{[DP: #1]}}
\newcommand{\onur}[1]{\textcolor{magenta}{[OK: #1]}}
\newcommand{\kaixuan}[1]{\textcolor{violet}{[KL: #1]}}
\else
\newcommand{\dimpp}[1]{}
\newcommand{\onur}[1]{}
\newcommand{\kaixuan}[1]{}
\fi

\newcommand{\mypar}[1]{\vspace{0mm}\noindent\textbf{#1}}

%% file: preamble.tex
\usepackage{dsfont}
\usepackage{mathtools}
\usepackage{multirow}
\usepackage[accsupp]{axessibility}  % Improves PDF readability for those with disabilities.

%% file: sec/0_abstract.tex
\begin{abstract}

Video Instance Segmentation (VIS) faces significant annotation challenges due to its dual requirements of pixel-level masks and temporal consistency labels. While recent unsupervised methods like VideoCutLER eliminate optical flow dependencies through synthetic data, they remain constrained by the synthetic-to-real domain gap. We present AutoQ-VIS, a novel unsupervised framework that bridges this gap through quality-guided self-training. Our approach establishes a closed-loop system between pseudo-label generation and automatic quality assessment, enabling progressive adaptation from synthetic to real videos. Experiments demonstrate state-of-the-art performance with 52.6 $\text{AP}_{50}$ on YouTubeVIS-2019 \texttt{val} set, surpassing the previous state-of-the-art VideoCutLER by 4.4$\%$, while requiring no human annotations. This demonstrates the viability of quality-aware self-training for unsupervised VIS.
%
% We will release the code and models upon acceptance.
The source code of our method is available at \href{https://github.com/wcbup/AutoQ-VIS}{here}.

\end{abstract}

%% file: sec/1_intro.tex
\section{Introduction}
\label{sec:intro}

\begin{figure}[t]
    \centering
    \includegraphics[width=1\linewidth]{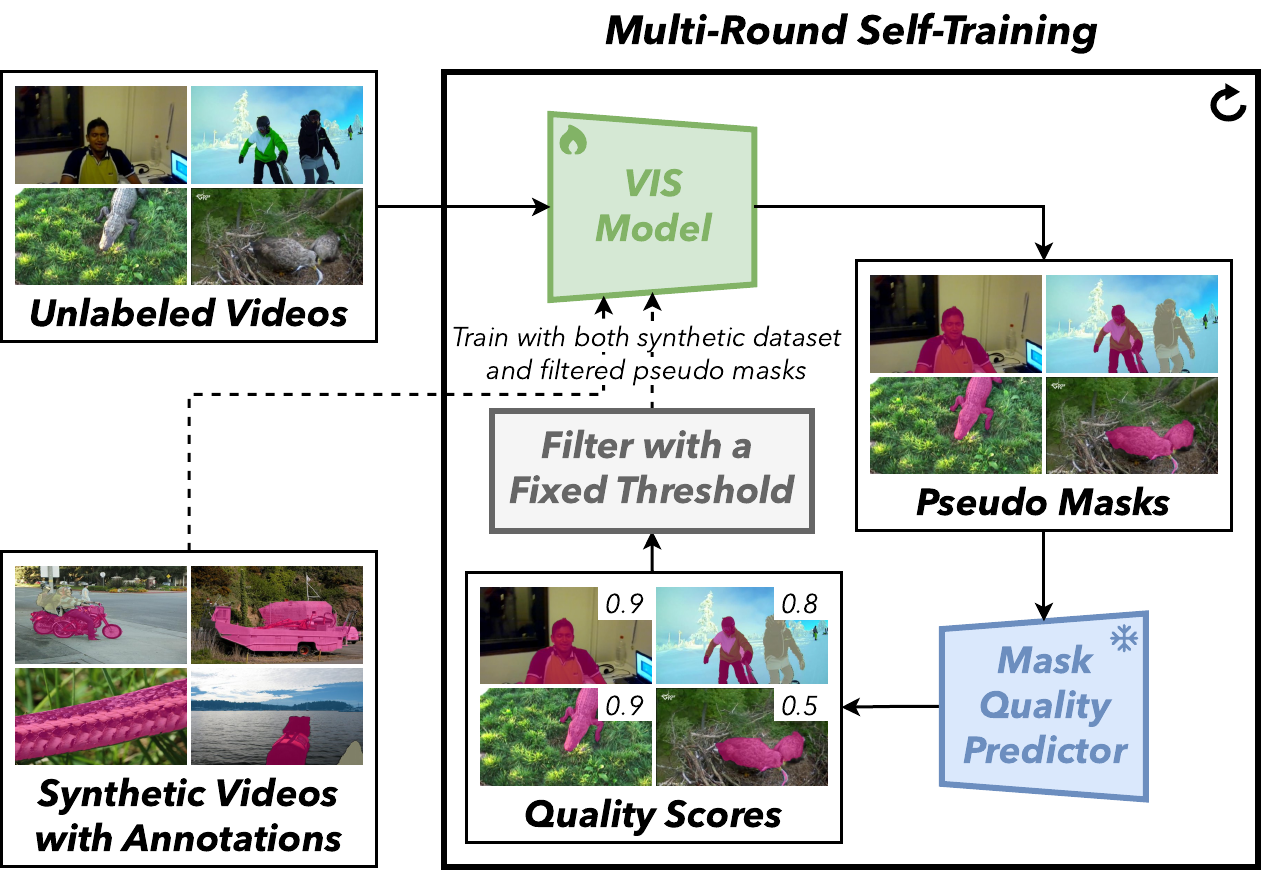}
    \caption{
        %The overall structure of AutoQ-VIS. The training dataset is initialized using the synthetic videos from VideoCutLER~\cite{videocutler}. In each round, we first train the VIS model on the established training dataset, then deploy it for pseudo-label generation on unlabeled video content. Following this, our quality predictor scores pseudo labels based on quality. Ultimately, we add the good-quality pseudo labels to the training dataset, which can be used for the next round of training. Through iterative training and dynamic dataset augmentation, we optimize the VIS model. The quality predictor is frozen during the self-training.
        %\textbf{Overview of the AutoQ-VIS pipeline.} We first initialize both the VIS model and the mask quality predictor by training on synthetic videos with annotations~\cite{videocutler} in the initial training stage. Then, in the multi-round self-training stage, at each round, the VIS model generates pseudo masks on unlabeled videos, which are then scored by the frozen quality predictor. Pseudo masks with high predicted quality scores are selected and added to the training set. The VIS model is retrained with both synthetic data and high-quality pseudo labels, enabling iterative self-training and progressive performance improvement.
        \textbf{AutoQ-VIS overview.} In the initial training stage, both the VIS model and the mask quality predictor are trained on synthetic videos with pseudo  annotations~\cite{videocutler}. During the multi-round self-training stage, the VIS model generates pseudo masks on unlabeled videos, which are then scored by the frozen quality predictor. Pseudo masks with high predicted quality are selected and added to the training set. The VIS model is subsequently retrained on both the synthetic data and the selected pseudo labels, enabling iterative refinement and progressive performance gains.
    }
    \label{fig:overall}
\end{figure}

Video Instance Segmentation (VIS) is the challenging task of simultaneously detecting, segmenting, and tracking object instances across video sequences~\cite{ytvis2019,wu2022seqformer,lin2020video,heo2022vita,wang2021end,zhang2023dvis}. This capability is fundamental for scene understanding in applications ranging from autonomous driving~\cite{driving_vis_dataset} to video content editing~\cite{vis_survey}. However, training high-performance VIS models typically requires pixel-level annotations across all frames~\cite{ytvis2019}. This process is expensive due to the labor-intensive nature of annotating temporal consistency and instance identities. As a result, there is an urgent need to develop unsupervised video instance segmentation frameworks that can accurately interpret video content and function effectively across diverse, unlabeled environments.
% To overcome this annotation bottleneck, we propose AutoQ-VIS, an unsupervised learning framework that eliminates manual labeling through iterative pseudo-label refinement with automatic quality assessment, enabling effective model training directly from unlabeled videos. \dimpp{this sentence does not fit here}
% ONUR: Follow this structure, separate paragraphs:
%1 Introduce the task with context and references.
%2 Highlight challenges and limitations of the state-of-the-art.
%3 Present your motivation: why this approach?
%4 Summarize the solution and its impact, with quantitative metrics

Prior work~\cite{motiongroup, emdriven, guessmove, bootstrapping, highly_probable_positive_features, transport_network, dense_video_seg} in unsupervised video segmentation predominantly addresses Video Object Segmentation (VOS), focusing on separating a single foreground object via motion or consistency cues. While OCLR~\cite{oclr} introduces unsupervised VIS that supports multiple instances, its predefined object count during training prevents dynamic adaptation to varying numbers of instances during inference. Furthermore, prior approaches~\cite{motiongroup, emdriven, guessmove, bootstrapping, oclr} rely on optical flow estimators (e.g.,  RAFT~\cite{raft}) that are trained on human-annotated datasets. VideoCutLER~\cite{videocutler} marks a breakthrough in unsupervised VIS and achieves unprecedented performance by demonstrating multi-instance segmentation without optical flow dependencies. Its core innovation lies in synthetic video generation via spatial augmentations of CutLER~\cite{cutler} pseudo-labels from ImageNet~\cite{imagenet}. However, VideoCutLER remains constrained by synthetic-to-real domain gaps and static instance modeling, i.e., the synthetic videos lack natural and realistic motion patterns. 
%\kaixuan{should we put mask scoring here? I can't find a smooth transition to put it here}
% ONUR: since we don't have a related work section, you can add more citations here

Building upon VideoCutLER's synthetic data paradigm, which generates training videos through spatial augmentations of static image pseudo-labels, we introduce AutoQ-VIS to address its critical domain gap limitation. While VideoCutLER's synthetic videos provide initial instance awareness, they lack natural motion patterns and real-world appearance variations, hindering adaptation to authentic video dynamics. Our framework bridges this synthetic-to-real domain gap through a self-training loop that progressively adds quality-filtered pseudo-labels from unlabeled real videos. Inspired by Mask Scoring R-CNN~\cite{maskscore}, which directly predicts mask quality scores via an auxiliary branch, we implement a quality assessment module for pseudo-label filtering of instance masks.
AutoQ-VIS advances unsupervised video instance segmentation through an iterative self-training paradigm with quality-aware pseudo-label selection (\cref{fig:overall}). The system initializes using synthetic video data from VideoCutLER, which provides pseudo-labels to bootstrap a VideoMask2Former~\cite{mask2former, videomask2former} VIS model and a specialized mask quality predictor (\cref{sec:initial_model}). The mask quality predictor 
%employs a Mask Scoring R-CNN-inspired architecture to 
estimates mask IoU quality scores by analyzing frame-level features and mask predictions. During multi-round optimization, the VIS model generates pseudo-labels on unlabeled videos, which are then scored by the quality predictor. High-quality pseudo-labels surpassing a fixed threshold are progressively incorporated into the training set, enabling dataset augmentation without any human supervision (\cref{sec:multi_round_training}). To enhance the mask head training, we employ a DropLoss that zeros out mask losses whose maximum ground‑truth overlap falls below 0.01 (\cref{sec:droploss}).  By alternating rounds of VIS training (with occasional weight resets) and quality‑based dataset expansion, AutoQ‑VIS dynamically enriches its training dataset and steadily sharpens segmentation performance. 

% \todo{mention uvo results}
AutoQ-VIS establishes a new state-of-the-art in unsupervised video instance segmentation, achieving 52.6\% $\text{AP}_{50}$ on the YouTubeVIS-2019 validation set~\cite{ytvis2019}. This represents a significant +4.4\% $\text{AP}_{50}$ improvement over the previous best-performing method VideoCutLER~\cite{videocutler}, demonstrating our framework's effectiveness in motion pattern understanding and instance-level discrimination. To further validate the approach's generalizability, we conduct additional evaluation on the UVO-Dense benchmark~\cite{uvo}, where our method achieves consistent performance gains (1.1\% $\text{AP}_{50}$ improvement over the previous state-of-the-art approaches) under dense object scenarios.

Our key contributions are threefold: \textbf{(1) Annotation-Free VIS Framework:} We propose AutoQ-VIS, an unsupervised framework that overcomes annotation dependency through cyclic pseudo-label refinement with automated quality control, enabling video instance segmentation training directly from unlabeled videos. \textbf{(2) Automatic Quality Assessment:} We propose a simple quality predictor that reliably filters pseudo labels across self-training rounds. \textbf{(3) New State-of-the-art Performance:} Our AutoQ-VIS achieves 52.6 ${\text{AP}_{50}}$ on YouTubeVIS-2019~\cite{ytvis2019} \texttt{val} split, surpassing the previous state-of-the-art VideoCutLER~\cite{videocutler} by 4.4 ${\text{AP}_{50}}$.

%% file: sec/2_relatedwork.tex
\section{Related Work}
\label{sec:related}

\textbf{Unsupervised video object segmentation (VOS)}~\cite{ytvos2018, davis16, segtrackv2, FBMS59, moca} targets identifying and segmenting moving objects as foreground elements in video sequences, generating pixel-level binary masks that distinguish these objects from background content without any human-annotated labels. The task applies to both scenarios, including those with a single object instance and those with multiple concurrent instances.
It is worth noting that some materials~\cite{lee2024guided, matnet, ren2021reciprocal, yang2021learning} refer to video salient object detection (SOD) as \textit{Unsupervised VOS}, which is a different task that requires human-annotated labels. Current unsupervised video segmentation methods predominantly rely on detection based on motion or consistency cues. Motion Group~\cite{motiongroup} uses optical flow to train an unsupervised video object segmentation network, and it uses only optical flow as input. Similar to Motion Group~\cite{motiongroup}, many methods~\cite{emdriven, bootstrapping, guessmove, yang2019unsupervised, oclr, ventura2019rvos} use optical flow as input or use it to supervise the model. However, they all rely on optical flow estimators (e.g., RAFT~\cite{raft}) that are trained on human-annotated datasets, and most of them~\cite{emdriven, bootstrapping, guessmove, yang2019unsupervised} can only track one object at a time. While some unsupervised video object segmentation methods~\cite{oclr, ventura2019rvos} support multi-object tracking, their multi-object tracking performance is very limited, which makes them unsuitable for the video instance segmentation task.

\noindent\textbf{Unsupervised video instance segmentation (VIS)}~\cite{ytvis2019, wu2022seqformer, lin2020video, heo2022vita, wang2021end, zhang2023dvis} targets identifying and segmenting moving objects from backgrounds and discriminating between distinct instances, all without any human-annotated labels.
VideoCutLER~\cite{videocutler} is the first unsupervised video segmentation method that is built for the VIS task, without using any human-annotated labels in the whole pipeline. VideoCutLER~\cite{videocutler} establishes a milestone in unsupervised video instance segmentation (VIS), surpassing previous benchmarks by demonstrating multi-instance segmentation without using optical flow. While its key innovation --- a synthetic video generation pipeline leveraging spatial augmentations of CutLER~\cite{cutler} pseudo-labels (originally derived from ImageNet~\cite{imagenet}) --- enables remarkable performance, the approach remains fundamentally limited by synthetic-to-real domain discrepancies and rigid instance modeling. Specifically, the static object configurations in synthesized videos fail to capture authentic motion dynamics observed in natural video sequences.

\noindent\textbf{Looped self-training method} is the method that uses the pseudo labels generated by the model to retrain the model~\cite{crst, zou2018unsupervised, noisy_student, st3d, twist, cutler}. This method is widely used in different areas related to unsupervised or semi-supervised training, for example, unsupervised domain adaptation~\cite{crst, zou2018unsupervised, st3d}, semi-supervised image classification~\cite{noisy_student}, semi-supervised 3D instance segmentation~\cite{twist}, unsupervised object detection and segmentation~\cite{cutler}, and human-in-the-loop systems for object detection~\cite{kuznetsova2020open,papadopoulos2016we}, instance segmentation~\cite{papadopoulos2021scaling,delatolas2024learning}, and
3D shape segmentation~\cite{yi2016scalable}.

%% file: sec/3_methodology.tex
\section{Method}

% We divide AutoQ‑VIS into three key stages: (1) \textbf{Initial training} (\cref{sec:initial_model}), where we jointly pretrain a VideoMask2Former VIS model and a Mask‑Scoring‑R‑CNN–style quality predictor on synthetic videos from VideoCutLER; (2) \textbf{Multi‑round self‑training} (\cref{sec:multi_round_training}), in which we alternate between: (i) using the VIS model to generate pseudo‑labels on unlabeled videos, (ii) scoring each pseudo‑label by regressing its IoU via the quality predictor, and (iii)
% adding only those pseudo-labels back into the training set that are both confident and have high predicted IoU, as determined by a combined quality score exceeding a fixed threshold;
% and (3) \textbf{DropLoss for the mask head} (\cref{sec:droploss}), which dynamically suppresses loss contributions from any predicted mask whose maximum ground‑truth overlap falls below a small IoU threshold and enhances the mask head training.  Through this iterative, quality‑guided dataset augmentation, our pipeline progressively refines instance segmentation accuracy without requiring any human annotations.

AutoQ-VIS operates through four parts: (1) \textbf{Initial Training} (\cref{sec:initial_model}): Jointly pretrain VideoMask2Former and the mask quality predictor on VideoCutLER's synthetic videos; (2) \textbf{Multi-Round Self-Training} (\cref{sec:multi_round_training}): Iteratively generate pseudo-labels on real unlabeled videos, filter via quality scores, and augment training data; (3) \textbf{DropLoss} (\cref{sec:droploss}): Suppress low-IoU mask predictions to enhance mask head training. This quality-guided pipeline progressively improves segmentation accuracy without any human annotations; (4) \textbf{Adaptive fusion}: Augment the dataset dynamically through adaptive fusion of newly curated annotations.

% \dimpp{we need an initial paragraph here as an overview that explains the pipeline!}

\begin{figure}[t]
    \centering
    \includegraphics[width=1\linewidth]{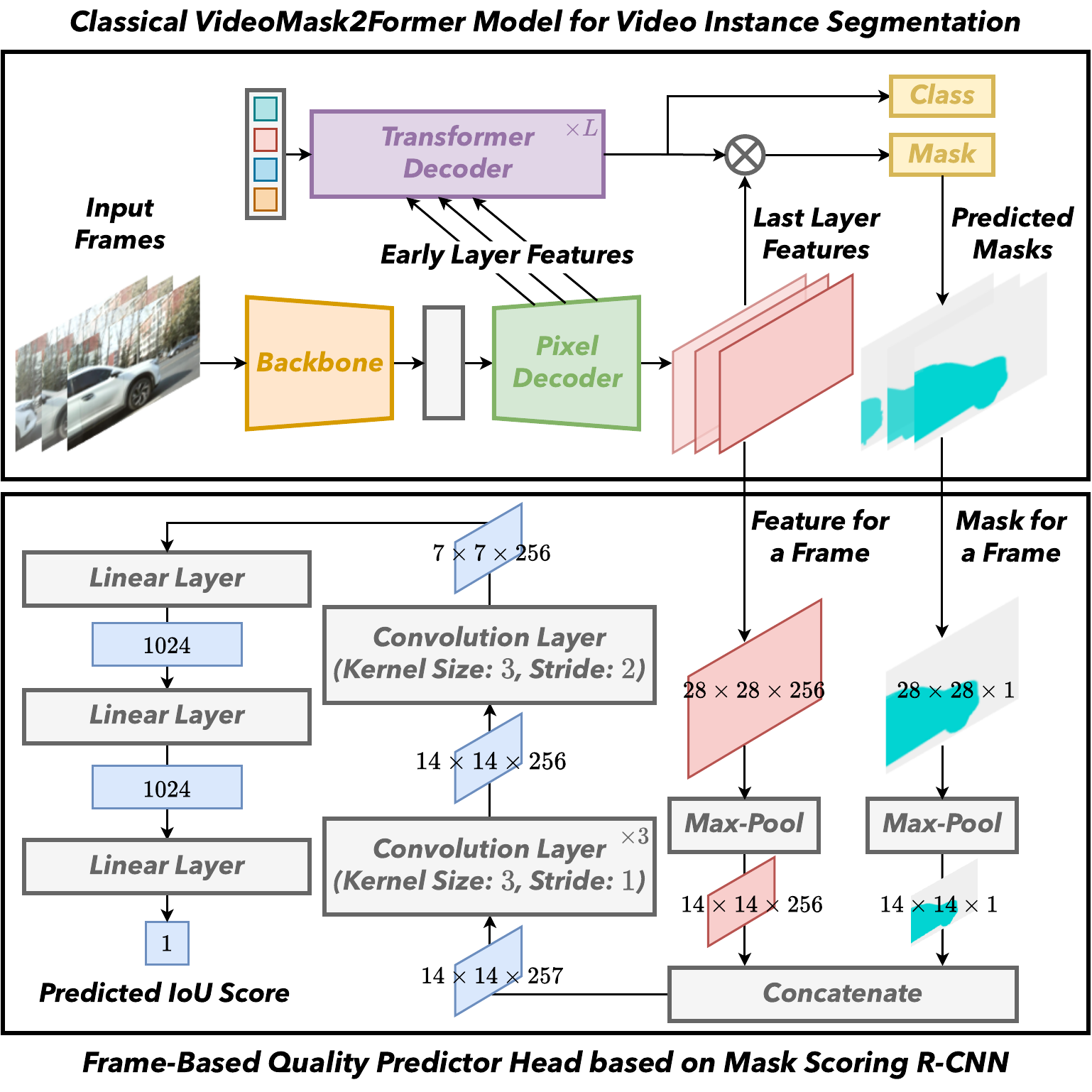}
    \caption{
    \textbf{Network architecture of VideoMask2Former~\cite{mask2former, videomask2former} and Mask Quality Predictor.} Our quality predictor integrates mask predictions and pixel decoder features following~\cite{maskscore}, employing a sequential architecture with four convolution layers (3$\times$3 kernels, final layer stride of 2 for spatial reduction) followed by three fully-connected layers that ultimately produce mask IoU predictions.}
    \label{fig:predictor}
\end{figure}

% ONUR: I did not like this title. Call it "Initialization stage"?
%\subsection{Initialize VIS model and quality predictor}
%\subsection{Initialization stage}
\subsection{Initial training stage}
\label{sec:initial_model}

% \paragraph{Video instance segmentation (VIS) model.}
\mypar{Video instance segmentation (VIS) model.}
Following VideoCutLER~\cite{videocutler}, we use the VideoMask2Former~\cite{mask2former, videomask2former} with the ResNet-50~\cite{resnet} backbone as our video instance segmentation (VIS) model.

% \paragraph{Quality predictor.}
\mypar{Quality predictor.}
For the quality predictor, as shown in~\cref{fig:predictor}, we use an architecture
%very
inspired by Mask Scoring R-CNN~\cite{maskscore}. Our architecture processes individual frame features and single-object mask predictions per inference step. Supervision is established through threshold-binarized (0.5) mask IoU between predictions and matched ground truths, optimized via $\ell_2$ regression loss.
\kaixuan{I add the following sentences to highlight the difference between our quality predictor and the mask scoring}
Although the architecture of our quality predictor and Mask Scoring R-CNN is similar, there is a key difference between our model and theirs. In Mask Scoring R-CNN, the input predicted mask is threshold-binarized (0.5). In our case, we find it crucial to use the raw predicted mask instead of the threshold-binarized predicted mask. If we use the threshold-binarized predicted mask, the quality predictor cannot produce any meaningful output. 

% \paragraph{Synthetic videos.} 
\mypar{Synthetic videos.} 
VideoCutLER~\cite{videocutler} provides a high-quality pseudo-labeled synthetic video dataset that was built on the unlabeled images from ImageNet~\cite{imagenet}, which is very suitable to train and initialize our VIS model and quality predictor. We also use the trained VideoMask2Former model~\cite{videomask2former} from VideoCutLER to initialize our VIS model.
% Instead of freezing the VIS model and training the quality predictor only, we found that training them both can lead to higher IoU prediction accuracy. 

\subsection{Multi-round self-training}
\label{sec:multi_round_training}

As shown in \cref{fig:overall}, we optimize the VIS model through iterative self-training and dynamic dataset augmentation. The training dataset is initialized using the synthetic videos from VideoCutLER~\cite{videocutler}. Empirically, we find that executing model parameter restoration from the initial model weight (trained in \cref{sec:initial_model}) achieves a better performance.

After training the VIS model, we use the predicted masks on unlabeled videos with a confidence score over $0.25$ as pseudo-labels. Then we use the quality predictor to predict the IoU of each pseudo-label predicted mask. Let $\hat{\text{IoU}}_l$ denote the predicted IoU of label $l$, and $s_l$ denotes the confidence score of label $l$ from the class head of the VIS model (see~\cref{fig:predictor}). We define the quality score of label $l$ as $Q_l = \hat{\text{IoU}}_l \cdot s_l$.
% \begin{equation}
%     Q_l = \hat{\text{IoU}}_l \cdot s_l
% \end{equation}

We implement quality-based pseudo-label selection using a fixed quality score threshold $\tau_{\text{th}}$. For each pseudo-label $l$, we select it if $Q_l \geq \tau_{th}$. At the end of each round, we add all the pseudo-labels to the training dataset.

% ONUR: This may be a paragraph, instead of a subsection????

\subsection{DropLoss for mask head}
\label{sec:droploss}

We enhance the mask head training by suppressing loss contributions from low-overlap predictions, following CutLER~\cite{cutler}. For each predicted mask $m_i$, we discard its loss contribution if its maximum ground truth IoU falls below the threshold $\tau^{\text{IoU}}$:
\begin{equation}
    \mathcal{L}_{\text{drop}}(m_i) = \mathds{1}(\text{IoU}_i^{\text{max}} > \tau^{\text{IoU}})\mathcal{L}_{\text{vanilla}}(m_i)
\end{equation}

\noindent Here, $\text{IoU}_i^{\text{max}}$ is the maximum overlap between $m_i$ and any ground truth mask, while $\mathcal{L}_{\text{vanilla}}$ denotes the original mask head loss from VideoMask2Former~\cite{mask2former, videomask2former}. We employ a low threshold ($\tau^{\text{IoU}} = 0.01$) to filter only near-zero overlap predictions.

\subsection{Adaptive fusion of new curated annotations and old curated annotations}

To augment the training dataset, at the end of each round of self-training, we need to merge the old curated annotations in the training dataset with the new curated annotations.

The augmentation process operates in two distinct modes depending on whether the video containing the new detections already exists in the training set. For novel videos, we perform bulk insertion of all qualified detections. For existing videos, we implement an instance-level fusion protocol that intelligently merges new predictions with old annotations while preserving temporal consistency.

Let $d_v$ denote an object detection for video $v$, formally:
\begin{equation}
    d_v = \{(S^t, m^t)\}_t^{T_v} 
\end{equation}
Here, $S^t \in \{0, 1\}$ indicates whether the object mask (pseudo label) of $d_v$ at frame $t$ is selected for self-training, $m^t$ denotes the binary object mask at frame $t$.

Let $\mathcal{T}^{(k)} = \{(v, \mathcal{D}_v)\}$ denote the training dataset at iteration $k$, where $\mathcal{D}_v$ contains all preserved detections (old annotations) for video $v$, and $\mathcal{D}_{\text{retained}}^{(k)}$ denote the set of detections that we retained after the filtering using the quality score threshold $\tau_{th}$ at the end of iteration $k$ (new detections). We need to merge the $\mathcal{D}_{\text{retained}}^{(k)}$ into the $\mathcal{T}^{(k)} = \{(v, \mathcal{D}_v)\}$. For each video, if it's not in the training dataset, we simply add all its detections to the training dataset. Formally, let $\mathcal{D}_{\text{retained},v}^{(k)}$ denote the detections belong to video $v$:
\begin{equation}
    \mathcal{D}_{\text{retained},v}^{(k)} = \left\{ d \in \mathcal{D}_{\text{retained}}^{(k)} \,\big|\, d \text{ belongs to video } v \right\}
\end{equation}

Let $\mathcal{V}_{\text{new}}^{(k)} = \pi_1(\mathcal{D}_{\text{retained}}^{(k)}) \setminus \pi_1(\mathcal{T}^{(k)})$ denote completely new videos, where $\pi_1$ projects to video identifiers. Let $\mathcal{T}_{\text{new}}^{(k)}$ denote labels of new videos:

\begin{equation}
\mathcal{T}_{\text{new}}^{(k+1)} =  \bigcup_{v \in \mathcal{V}_{\text{new}}^{(k)}} \left\{ \left(v, \mathcal{D}_{\text{retained},v}^{(k)} \right) \right\}
\end{equation}

For videos already present in the training set, we implement a temporal-aware fusion protocol that resolves conflicts between new detections and old annotations.

To merge the existing detections and new detections, we need to identify whether two detections overlap. We define the spatiotemporal overlap predicate for two detections:
\begin{equation}
    \phi(d_{\text{new}, v}, d_{\text{exist}, v}) \triangleq \exists t \in [1, T_v]: \frac{\|m_{\text{new}}^t \cap m_{\text{exist}}^t\|}{\|m_{\text{new}}^t \cup m_{\text{exist}}^t\|} \geq 0.5
\end{equation}

% This predicate is very similar to what we use in spatiotemporal NMS.
If in any frames, two detections' masks overlap (IoU $\geq$ 0.5), we consider them overlapping detections. If two detections overlap, we need to fuse them. We fuse the new detection and the existing detection frame by frame. For one frame, if only the existing detection's label is selected, we use its label; otherwise, we use the new detection's label. Let $\mathcal{F}$ denote the fusion operation:
\begin{equation}
    \mathcal{F}(d_{\text{new},v}, d_{\text{exist},v}) =  \{\mathcal{S}_{\text{merge}}^t, m_{\text{merge}}^t\}_{t=1}^{T_v}
\end{equation}
where:
\begin{equation}
    \mathcal{S}_{\text{merge}}^t = \max(\mathcal{S}_{\text{new}}^t, \mathcal{S}_{\text{exist}}^t)
\end{equation}
\begin{equation}
    m_{\text{merge}}^t = \begin{cases} 
m_{\text{exist}}^t & \text{if } \mathcal{S}_{\text{exist}}^t = 1 \land \mathcal{S}_{\text{new}}^t = 0 \\
m_{\text{new}}^t & \text{otherwise}
\end{cases}
\end{equation}

If the new detection does not overlap with any existing detections, we simply add it. Let $\mathcal{V}_{\text{exist}}^{(k)} = \pi_1(\mathcal{T}^{(k)})$ denote existing videos. For each existing video $v \in \mathcal{V}_{\text{exist}}^{(k)}$, we process all new detections $\mathcal{D}_{\text{retained},v}^{(k)}$ through:
\begin{equation}
    \mathcal{D}_v^{(k+1)} = \mathcal{D}_v^{(k)} \bigoplus \mathcal{D}_{\text{retained},v}^{(k)}
\end{equation}
where the operation $\bigoplus$ is defined as:
\begin{equation}
    \mathcal{D} \bigoplus \mathcal{D}' = \bigcup_{d' \in \mathcal{D}'} \left( \mathcal{D} \oplus d' \right)
\end{equation}
with per-detection fusion:
\begin{equation}
\mathcal{D} \oplus d_{\text{new}} = 
\begin{cases}
\mathcal{D} \cup \{d_{\text{new}}\} & \text{if } \forall d_{\text{exist}} \in \mathcal{D}, \\
& \quad \neg\phi(d_{\text{new}}, d_{\text{exist}}) \\
\mathcal{D} \setminus d_{\text{exist}} \cup \mathcal{F}(d_{\text{new}}, d_{\text{exist}}) & \text{if } \exists d_{\text{exist}} \in \mathcal{D}, \\
& \quad \phi(d_{\text{new}}, d_{\text{exist}})
\end{cases}
\end{equation}

In the end, we merge the labels of new videos and existing videos:
\begin{equation}
    \mathcal{T}^{(k+1)} = \mathcal{T}_{\text{new}}^{(k+1)} \cup \mathcal{T}_{\text{exist}}^{(k+1)}
\end{equation}
where:
\begin{equation}
    \mathcal{T}_{\text{exist}}^{(k+1)} =  \bigcup_{v \in \mathcal{V}_{\text{exist}}^{(k)}} \left\{ \left(v, \mathcal{D}_{v}^{(k+1)} \right) \right\}
\end{equation}

During training, for each video, we need to sample three frames as the input of the model (following VideoCutLER~\cite{videocutler}). To provide good-quality labels, we only sample those frames where all their detections are selected. For each video $v \in \mathcal{V}_{\text{train}}^{(k+1)}$, we define the eligible frame set:
\begin{equation}
\mathcal{F}_{\text{eligible}}^{(k+1)}(v) = \left\{ t \in [1,T_v] \,\big|\, \forall d \in \mathcal{D}_v^{(k+1)}: \mathcal{S}_d^t = 1 \right\}
\end{equation}

The training batch is constructed through uniform sampling:
\begin{equation}
\mathcal{B}_{\text{train}}^{(k+1)}(v) = \mathop{\mathrm{UniformSample}}\limits_{3} \left( \mathcal{F}_{\text{eligible}}^{(k+1)}(v) \right)
\end{equation}

Although we only use a subset of pseudo-labels, all pseudo-labels persist in the training set regardless of selection status, enabling progressive refinement.

% To balance data distribution between VideoCutLER's~\cite{videocutler} extensive synthetic videos and our pseudo-labeled videos, we implement a balanced stochastic sampling strategy. Each training batch has an equal probability (50\%) of being drawn from either the synthetic videos or the pseudo-labeled set. This prevents overfitting to either domain while maintaining the synthetic data's regularization benefits during self-training. \todo{move it to implement detail}

% ONUR: No need for the YoutubeVIS row, always the same eval set
\begin{table}[!t]
\resizebox{\columnwidth}{!}{%
\begin{tabular}{lrrrrrrr}
\toprule

\multicolumn{1}{c}{}                                 & \multicolumn{7}{c}{\textbf{YouTubeVIS-2019}}   \\  \cline{2-8}
\multicolumn{1}{c}{\multirow{-2}{*}{\textbf{Method}}} &
  \multicolumn{1}{l}{\textbf{${\text{AP}}_{50}$}} &
  \multicolumn{1}{l}{\textbf{${\text{AP}}_{75}$}} &
  \multicolumn{1}{l}{\textbf{${\text{AP}}$}} &
  \multicolumn{1}{l}{\textbf{${\text{AP}}_{S}$}} &
  \multicolumn{1}{l}{\textbf{${\text{AP}}_{M}$}} &
  \multicolumn{1}{l}{\textbf{${\text{AP}}_{L}$}} &
  \multicolumn{1}{l}{\textbf{${\text{AR}}_{10}$}} \\ \midrule

% \textbf{Method} & 
% \textbf{${\text{AP}}_{50}$} &
% \textbf{${\text{AP}}_{75}$} &
% \textbf{${\text{AP}}$} & 
% \textbf{${\text{AP}}_{S}$} & 
% \textbf{${\text{AP}}_{M}$} & 
% \textbf{${\text{AP}}_{L}$} & 
% \textbf{${\text{AR}}_{10}$} \\ \midrule

MotionGroup~\cite{motiongroup} & 0.5  & 0.0  & 0.1  & 0.0 & 0.4  & 0.1  & 1.2  \\
OCLR~\cite{oclr}               & 5.5  & 0.3  & 1.6  & 0.1 & 1.6  & 6.1  & 11.5 \\
CutLER~\cite{cutler}           & 36.4 & 13.5 & 16.0 & 3.5 & 13.9 & 26.0 & 29.8 \\
VideoCutLER~\cite{videocutler} & 48.2 & 22.9 & 24.5 & 6.7 & 17.7 & 36.3 & 42.3 \\
AutoQ-VIS                                           & 52.6 & 28.2 & 28.1 & 6.7 & 21.2 & 40.7 & 42.5 \\
\textit{vs. previous SOTA} &
  {\color[HTML]{009901} +4.4} &
  {\color[HTML]{009901} +5.3} &
  {\color[HTML]{009901} +3.6} &
  {\color[HTML]{009901} +0.0} &
  {\color[HTML]{009901} +3.5} &
  {\color[HTML]{009901} +4.4} &
  {\color[HTML]{009901} +0.2} \\

\bottomrule
\end{tabular}%
}
\caption{
\textbf{YouTubeVIS-2019 \texttt{val}.} We reproduced MotionGroup~\cite{motiongroup}, OCLR~\cite{oclr}, CutLER~\cite{cutler}, and VideoCutLER~\cite{videocutler} results with the official code and checkpoints. AutoQ-VIS outperforms the state-of-the-art VideoCutLER by 4.4 $\text{AP}_{50}$. We evaluate results on YouTubeVIS-2019's \texttt{val} split in a class-agnostic manner.
}
\label{tab:ytvis19val}
\end{table}

%% file: sec/4_experiment.tex
\section{Experiments}

%\todo{Initial paragraph}

%\subsection{Implementation details}

% \paragraph{Datasets.}
\mypar{Datasets.}
Our model is trained on synthetic videos from VideoCutLER~\cite{videocutler} and the unlabeled \texttt{train} split of YouTubeVIS-2019~\cite{ytvis2019}. We evaluate our model's performance on the \texttt{val} split of YouTubeVIS-2019 in a class-agnostic manner. YouTubeVIS-2019 contains 2,883 high-resolution YouTube videos, annotated at 6 FPS (2,238 training videos, 302 validation videos, and 343 test videos).

\kaixuan{We don't need to describe UVO-Dense here, right?}

% \paragraph{Evaluation metrics.}
\mypar{Evaluation metrics.}
Following VideoCutLER~\cite{videocutler}, we use Average Precision (AP) and Average Recall (AR) as evaluation metrics. We evaluate the models in a class-agnostic manner, treating all classes as a single one during evaluation.

% \paragraph{Implementation details.}
\mypar{Implementation details.}
For the initial training stage, we use the pretrained VideoMask2Former model~\cite{mask2former, videomask2former} from VideoCutLER~\cite{videocutler} to initialize our VIS model. Then we train the VIS model and quality predictor on synthetic videos from VideoCutLER for 8,000 iterations using a single V100 GPU, with a batch size of 2 and a learning rate of $2 \times 10^{-5}$. For each round of multi-round self-training, we train the VIS model for 10,000 iterations on two V100 GPUs, with a batch size of 4 and a learning rate of $2 \times 10^{-5}$. In practice, we find that two rounds of self-training and a quality score threshold $\tau_{th}$ of 0.75 provide the best performance. 

To balance data distribution between VideoCutLER's~\cite{videocutler} extensive synthetic videos and our pseudo-labeled videos, we implement a balanced stochastic sampling strategy. Each training batch has an equal probability (50\%) of being drawn from either the synthetic videos or the pseudo-labeled set. This prevents overfitting to either domain while maintaining the synthetic data's regularization benefits during self-training. %\todo{move it to implement detail}

% We adopt a balanced stochastic sampling strategy, drawing each training batch with equal probability (50\%) from VideoCutLER's synthetic videos~\cite{videocutler} or pseudo-labeled data. This mitigates domain overfitting while preserving the synthetic data’s regularization benefits in self-training.

\subsection{Experimental results}

% \paragraph{Comparison with the-state-of-the-art method.}
\mypar{Comparison with the-state-of-the-art method.}
We compare AutoQ-VIS with previous top-performing methods on YouTubeVIS-2019~\cite{ytvis2019} in~\cref{tab:ytvis19val}. AutoQ-VIS achieves a remarkable improvement (about 4.4\% $\text{AP}_{50}$). Especially for $\text{AP}_{75}$, AutoQ-VIS can outperform the state-of-the-art VideoCutLER~\cite{videocutler} by 5.3\%. The observed improvements of +2.4\% $\text{AP}_{L}$, +1.2\% $\text{AP}_{M}$, and +0.0\% $\text{AP}_{L}$ suggest that our pipeline primarily enhances segmentation performance for medium and large objects, with negligible gains on other categories.

% To further evaluate the effectiveness of our pipeline, we also compare AutoQ-VIS with the state-of-the-art VideoCutLER~\cite{videocutler} on more challenging UVO-Dense~\cite{uvo} in~\cref{tab:uvo_dense}. AutoQ-VIS demonstrates a performance gain of 1.1\% in $\text{AP}_{50}$, establishing measurable improvements in video instance segmentation accuracy. \todo{mention our change to the quality score threshold, put this in the end}

\begin{table}[!t]
\resizebox{\columnwidth}{!}{%
\begin{tabular}{lrrrrrrr}
\toprule

\multicolumn{1}{c}{}                                 & \multicolumn{7}{c}{\textbf{YouTubeVIS-2019}}   \\  \cline{2-8}
\multicolumn{1}{c}{\multirow{-2}{*}{\textbf{Method}}} &
  \multicolumn{1}{l}{\textbf{${\text{AP}}_{50}$}} &
  \multicolumn{1}{l}{\textbf{${\text{AP}}_{75}$}} &
  \multicolumn{1}{l}{\textbf{${\text{AP}}$}} &
  \multicolumn{1}{l}{\textbf{${\text{AP}}_{S}$}} &
  \multicolumn{1}{l}{\textbf{${\text{AP}}_{M}$}} &
  \multicolumn{1}{l}{\textbf{${\text{AP}}_{L}$}} &
  \multicolumn{1}{l}{\textbf{${\text{AR}}_{10}$}} \\ \midrule

% \textbf{Method} & 
% \textbf{${\text{AP}}_{50}$} &
% \textbf{${\text{AP}}_{75}$} &
% \textbf{${\text{AP}}$} & 
% \textbf{${\text{AP}}_{S}$} & 
% \textbf{${\text{AP}}_{M}$} & 
% \textbf{${\text{AP}}_{L}$} & 
% \textbf{${\text{AR}}_{10}$} \\ \midrule

Theoretical limit                                    & 76.8 & 48.7 & 46.8 & 13.5 & 43.6 & 62.9 & 58.0 \\
Practical limit                                & 62.7 & 33.2 & 33.9 &  4.3 & 27.3 &  53.2 & 47.5   \\
AutoQ-VIS                                           & 52.6 & 28.2 & 28.1 & 6.7  & 21.2 & 40.7 & 42.5 \\ 
\bottomrule
\end{tabular}%
}
\caption{\textbf{Comparison with the theoretical and practical limit.} 
\textit{Theoretical Limit}: Upper-bound performance achieved by training on ground-truth labels from YouTubeVIS-2019 \texttt{train} split in class-agnostic mode, representing ideal supervision conditions.  
\textit{Practical Limit}: Best attainable performance when using all pseudo-labels with $\text{IoU}\geq0.5$ against ground truth, simulating perfect pseudo-label selection.
% \textit{Theoretical limit}: Upper-bound benchmark obtained by training on ground-truth labels from YouTubeVIS-2019 \texttt{train} split in class-agnostic manner. \textit{Practical limit}: Performance ceiling when incorporating all pseudo labels having $\text{IoU}\geq0.5$ with ground truth.
}
\label{tab:limit_comparison}
\end{table}

% \paragraph{Comparison with the theoretical and practical limit.}
\mypar{Comparison with the theoretical and practical limit.}
\cref{tab:limit_comparison} reveals a significant performance gap (10.1 $\text{AP}_{50}$) between AutoQ-VIS and the practical upper bound, indicating substantial potential for improvement through enhanced pseudo-label utilization. The theoretical limit represents a fully supervised training using all ground-truth annotations from YouTubeVIS-2019
\texttt{train} set. The practical limit represents an oracle experiment that simulates perfect pseudo-label selection by using all predictions with $\text{IoU} \geq 0.5$ against ground truth.

% with the best attainable performance when using all pseudo-labels with $\text{IoU}\geq0.5$ against ground truth, simulating perfect pseudo-label selection.

\begin{figure*}[ht]
    \centering
    \includegraphics[width=1\linewidth]{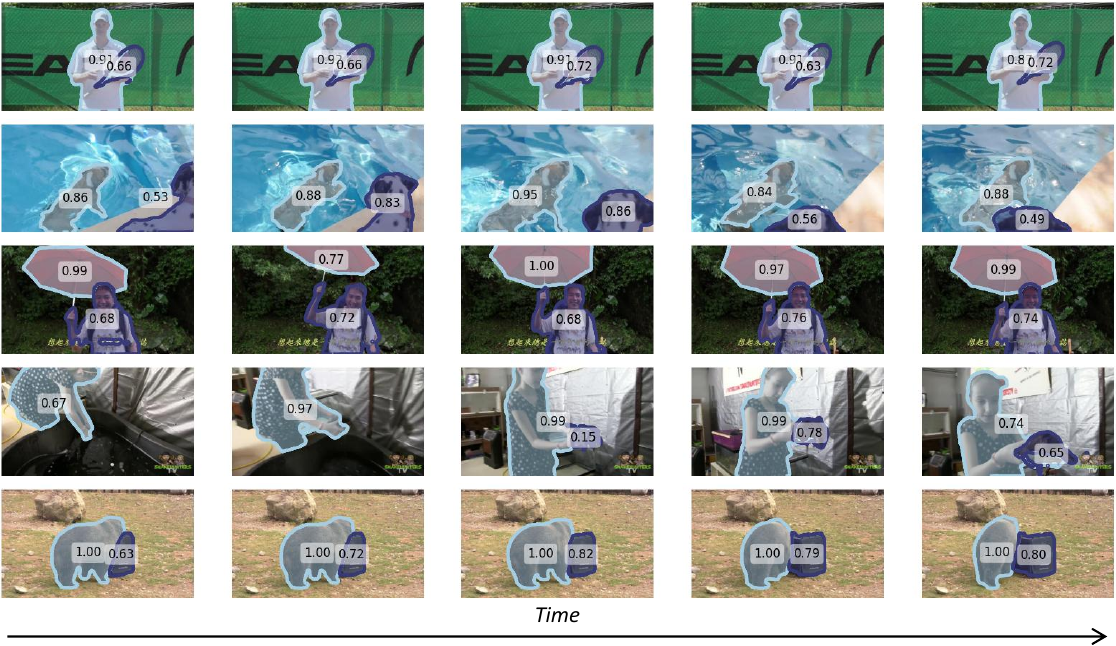}
    \caption{
        \textbf{The qualitative results of AutoQ-VIS on YouTubeVIS-2019 \texttt{val} split.} The quality scores are shown in the center of each object. The visual results demonstrate AutoQ-VIS' proficiency in simultaneous multi-instance segmentation, persistent object tracking, and per-mask quality assessment across video sequences. %\todo{replace with val results}
    }
    \label{fig:big_quali_res}
\end{figure*}

\begin{figure}[ht]
    \centering
    \includegraphics[width=1\linewidth]{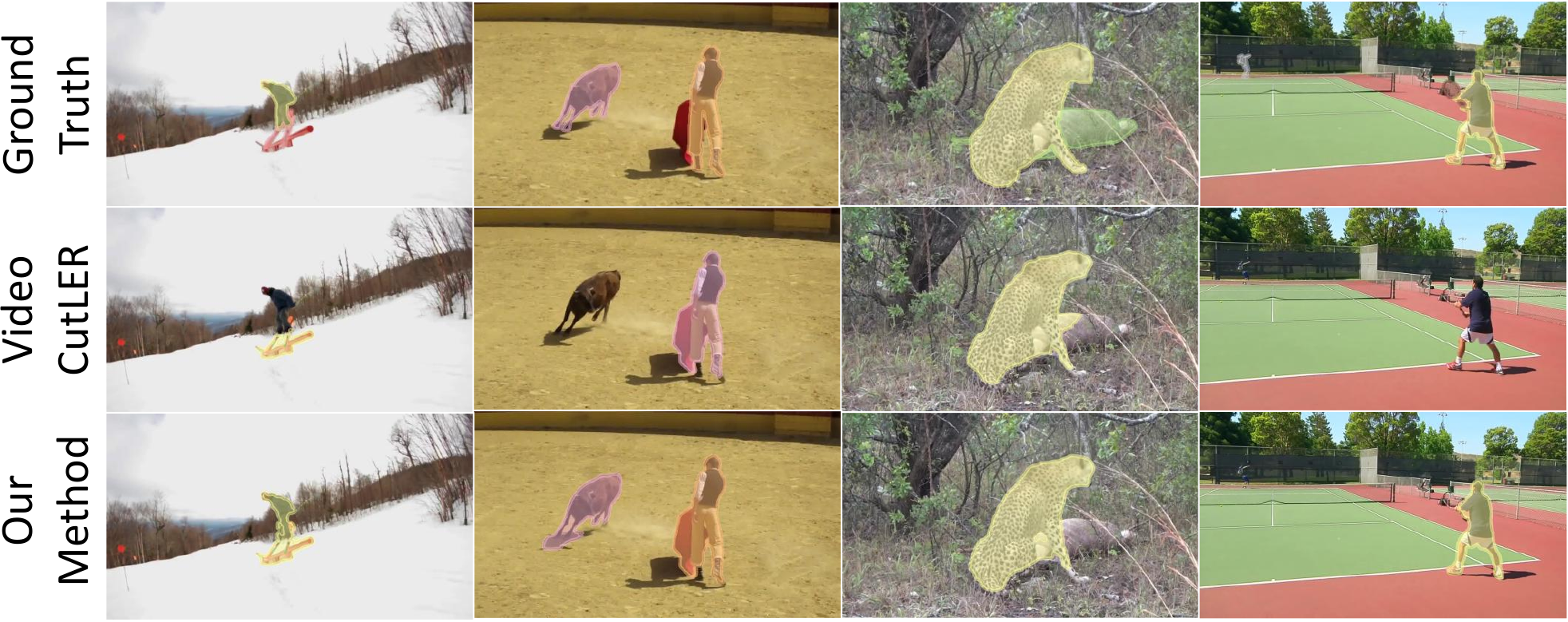}
    \caption{\textbf{The qualitative comparison on YouTubeVIS-2019 \texttt{val} split.} AutoQ-VIS demonstrates superior instance discovery capabilities compared to VideoCutLER~\cite{videocutler}: (1) Enhanced multi-object detection capacity, particularly for semantically distinct instances (e.g., person and bull in Column 2); (2) Improved segmentation fidelity through precise boundary delineation (e.g., the leopard in Column 3). (3) Better comprehensive instance coverage, eliminating false negatives (e.g., detecting humans in Columns 1 \& 4 that VideoCutLER completely misses, even without occlusion or scale challenges). 
    % \todo{take some figures from fig 8 make it two column}
    }
    % ONUR: I dont want to add limitations, bad examples to this figure. These are our only visual samples
    \label{fig:qualitative_results}
\end{figure}

% ONUR: we can merge quant and qualitative results?
% \paragraph{Qualitative results.}
\mypar{Qualitative results.}
Qualitative results in~\cref{fig:big_quali_res} illustrate AutoQ-VIS's capability in integrated multi-object segmentation, temporally consistent tracking, and mask-level quality prediction throughout video sequences. Fig.~\ref{fig:qualitative_results} demonstrates AutoQ-VIS's advancements over VideoCutLER~\cite{videocutler}. As we observe, AutoQ-VIS is capable of discovering more objects and producing higher-quality segmentation masks.

\subsection{Ablation studies}

\begin{table}[t]
\resizebox{\columnwidth}{!}{%
\begin{tabular}{lrrrrrrr}
\toprule

\multicolumn{1}{c}{}                                 & \multicolumn{7}{c}{\textbf{YouTubeVIS-2019}}   \\  \cline{2-8}
\multicolumn{1}{c}{\multirow{-2}{*}{\textbf{Method}}} &
  \multicolumn{1}{l}{\textbf{${\text{AP}}_{50}$}} &
  \multicolumn{1}{l}{\textbf{${\text{AP}}_{75}$}} &
  \multicolumn{1}{l}{\textbf{${\text{AP}}$}} &
  \multicolumn{1}{l}{\textbf{${\text{AP}}_{S}$}} &
  \multicolumn{1}{l}{\textbf{${\text{AP}}_{M}$}} &
  \multicolumn{1}{l}{\textbf{${\text{AP}}_{L}$}} &
  \multicolumn{1}{l}{\textbf{${\text{AR}}_{10}$}} \\ \midrule

% \textbf{Method} & 
% \textbf{${\text{AP}}_{50}$} &
% \textbf{${\text{AP}}_{75}$} &
% \textbf{${\text{AP}}$} & 
% \textbf{${\text{AP}}_{S}$} & 
% \textbf{${\text{AP}}_{M}$} & 
% \textbf{${\text{AP}}_{L}$} & 
% \textbf{${\text{AR}}_{10}$} \\ \midrule

w/o quality predictor                            & 50.5 & 25.9 & 27.2 & 5.7 & 19.0 & 40.5 & \textbf{43.3} \\
w/o DropLoss                                     & 48.0 & 23.7 & 24.6 & 3.8 & 16.9 & 37.8 & 39.2 \\
w/o resetting each round                         & 51.6 & 28.0 & \textbf{28.2} & 6.4 & \textbf{22.8} & 40.6 & 42.9 \\
w/o adaptive fusion                              & 49.7 & 26.5 & 26.8 & 6.1 & 18.9 & 40.1 & 41.4 \\
w/o freezing predictor                           & 49.2 & 25.8 & 26.5 & \textbf{6.9} & 18.8 & 39.3 & 40.9 \\
AutoQ-VIS                                        & \textbf{52.6} & \textbf{28.2} & 28.1 & 6.7 & 21.2 & \textbf{40.7} & 42.5 \\ 
VideoCutLER~\cite{videocutler}                   & 48.2 & 22.9 & 24.5 & 6.7 & 17.7 & 36.3 & 42.3 \\
\bottomrule
\end{tabular}%
}
\caption{
\textbf{Ablation study on the contribution of each component.} \textit{Without quality predictor:} We remove the quality predictor, and use the confidence score $s_l$ from the class head of VIS model (see~\cref{fig:predictor}) as quality score $Q_l$ with threshold $\tau_{th} = 0.85$. \textit{Without DropLoss:} We use the vanilla loss for the mask head instead of the DropLoss. \textit{Without resetting each round:} The model weights are not reset at the beginning of each round. \textit{Without Adaptive fusion:} Instead of fusing the newly curated annotations and old curated annotations, we simply replace the old curated annotations in the training dataset with new curated annotations. \textit{Without freezing predictor:} We do not freeze the weight of the quality predictor and train it together with the VIS model. %\todo{discuss the new results, show the videocutler here. which has huge impact, which doesn't}
}
\label{tab:component_ablation}
\end{table}

% \paragraph{Component ablation analysis.}
\mypar{Component ablation analysis.}
\cref{tab:component_ablation} quantifies individual component contributions through progressive additions. The DropLoss contributes the most significant performance improvement (+4.6\% $\text{AP}_{50}$) in our framework. While originally developed by CutLER~\cite{cutler} for image-level segmentation training, this technique was notably absent in VideoCutLER's~\cite{videocutler} video instance segmentation pipeline. Our quantitative results in~\cref{tab:component_ablation} demonstrate the importance of reintroducing DropLoss in our video-domain self-training pipeline.

Subsequent components demonstrate incremental gains: freezing the quality predictor contributes +3.4\% $\text{AP}_{50}$, adaptive fusion adds +2.9\% $\text{AP}_{50}$, and the quality predictor itself provides +2.1\% $\text{AP}_{50}$ improvement. 
Freezing quality predictor prevents overestimation bias by decoupling the optimization process between the VIS model and the quality predictor. Without parameter freezing, both components would reinforce pseudo-label errors through mutual confirmation during self-training cycles, thereby inducing progressively amplified confidence miscalibration in successive iterations. \kaixuan{Is this explanation reasonable?}
Adaptive fusion of new curated annotations and old curated annotations helps the model achieve progressive improvements.
Notably, even the confidence score baseline surpasses VideoCutLER by +2.3 $\text{AP}_{50}$, demonstrating fundamental advantages of our self-training method. While model resetting yields marginal gains (+1.0 $\text{AP}_{50}$, +0.2 $\text{AP}_{75}$), it maintains baseline $\text{AP}$ performance.  
%\todo{add freezing predictor}

\begin{table}[t]
\resizebox{\columnwidth}{!}{%
\begin{tabular}{lrrrrrrr}
\toprule

\multicolumn{1}{c}{}                                 & \multicolumn{7}{c}{\textbf{YouTubeVIS-2019}}   \\  \cline{2-8}
\multicolumn{1}{c}{\multirow{-2}{*}{\textbf{Self-training}}} &
  \multicolumn{1}{l}{\textbf{${\text{AP}}_{50}$}} &
  \multicolumn{1}{l}{\textbf{${\text{AP}}_{75}$}} &
  \multicolumn{1}{l}{\textbf{${\text{AP}}$}} &
  \multicolumn{1}{l}{\textbf{${\text{AP}}_{S}$}} &
  \multicolumn{1}{l}{\textbf{${\text{AP}}_{M}$}} &
  \multicolumn{1}{l}{\textbf{${\text{AP}}_{L}$}} &
  \multicolumn{1}{l}{\textbf{${\text{AR}}_{10}$}} \\ \midrule

% \textbf{Self-training} & 
% \textbf{${\text{AP}}_{50}$} &
% \textbf{${\text{AP}}_{75}$} &
% \textbf{${\text{AP}}$} & 
% \textbf{${\text{AP}}_{S}$} & 
% \textbf{${\text{AP}}_{M}$} & 
% \textbf{${\text{AP}}_{L}$} & 
% \textbf{${\text{AR}}_{10}$} \\ \midrule

1 round                                                                      & 51.3 & 26.5 & 26.9 & \textbf{7.0} & \textbf{22.3} & 38.4 & \textbf{43.2} \\
2 rounds                                                                      & \textbf{52.6} & \textbf{28.2} & \textbf{28.1} & 6.7 & 21.2 & \textbf{40.7} & 42.5 \\
3 rounds                                                                      & 52.0 & 27.0 & 27.7 & 6.2 & 21.8 & 40.1 & 42.1 \\
\bottomrule
\end{tabular}%
}
\caption{
\textbf{Ablation study on the number of self-training rounds.} Our framework achieves peak performance (52.6 $\text{AP}_{50}$) at the second round before gradual degradation (-0.6 $\text{AP}_{50}$) from pseudo-label noise accumulation. This establishes round 2 as the optimal stopping point to balance accuracy and error propagation risks.
}
\label{tab:round_num}
\end{table}

% \begin{figure}[t]
%     \centering
%     \includegraphics[width=1\linewidth]{figures/area_distribution_all_rounds_with_gt.pdf}
%     \caption{
%         \textbf{Comparison of the distributions of object areas of the ground truth labels and the pseudo labels at round 2.} Here, \textit{Log Area} denotes the natural logarithm of the normalized area (object area divided by image area). Compared to the ground truth labels, the pseudo labels contain more large objects. \kaixuan{Where should I put the analysis of this figure?}
%     }
%     \label{fig:pseudo_labels_and_gt_area}
% \end{figure}

% \paragraph{Self-training round analysis.}
\mypar{Self-training round analysis.}
\cref{tab:round_num} tests how many self-training rounds work best. The model hits its peak (52.6 $\text{AP}_{50}$) at round 2, then slowly gets worse. This drop occurs because, as we perform more rounds, mistakes in the pseudo-labels pile up.
% Based on these results, we recommend stopping after 2 rounds to get the best performance before these errors hurt the model.

\begin{table}[t]
\resizebox{\columnwidth}{!}{%
\begin{tabular}{rrrrrrrr}
\toprule

\multicolumn{1}{c}{}                                 & \multicolumn{7}{c}{\textbf{YouTubeVIS-2019}}   \\  \cline{2-8}
\multicolumn{1}{c}{\multirow{-2}{*}{\textbf{$\tau_{th}$}}} &
  \multicolumn{1}{l}{\textbf{${\text{AP}}_{50}$}} &
  \multicolumn{1}{l}{\textbf{${\text{AP}}_{75}$}} &
  \multicolumn{1}{l}{\textbf{${\text{AP}}$}} &
  \multicolumn{1}{l}{\textbf{${\text{AP}}_{S}$}} &
  \multicolumn{1}{l}{\textbf{${\text{AP}}_{M}$}} &
  \multicolumn{1}{l}{\textbf{${\text{AP}}_{L}$}} &
  \multicolumn{1}{l}{\textbf{${\text{AR}}_{10}$}} \\ \midrule

% \textbf{$\tau_{th}$} & 
% \textbf{${\text{AP}}_{50}$} &
% \textbf{${\text{AP}}_{75}$} &
% \textbf{${\text{AP}}$} & 
% \textbf{${\text{AP}}_{S}$} & 
% \textbf{${\text{AP}}_{M}$} & 
% \textbf{${\text{AP}}_{L}$} & 
% \textbf{${\text{AR}}_{10}$} \\ \midrule

0.95 & 48.7 & 25.3 & 26.1 & 5.9 & 18.8 & 38.5 & 40.7 \\
0.85 & 48.8 & 24.6 & 26.0 & 5.8 & 18.6 & 38.6 & 41.2 \\
0.75 & \textbf{52.6} & \textbf{28.2} & \textbf{28.1} & \textbf{6.7} & \textbf{21.2} & \textbf{40.7} & \textbf{42.5} \\
0.50  & 52.4 & 25.7 & 27.1 & 6.5 & 20.1 & 39.9 & 42.2 \\ \bottomrule
\end{tabular}%
}
\caption{
\textbf{Ablation study on the quality score threshold $\tau_{th}$.} Optimal performance (52.6 $\text{AP}_{50}$) emerges at $\tau_{th}=0.75$, balancing valid sample retention and noise suppression. Lower threshold ($\tau_{th}=0.50$) degrades results by admitting too many low-quality predictions, while the higher thresholds ($\tau_{th}=0.95$ and $\tau_{th}=0.80$) oversuppress valid samples.
}
\label{tab:quality_threshold}
\end{table}

% \paragraph{Quality score threshold $\tau_{th}$ analysis.}
\mypar{Quality score threshold $\tau_{th}$ analysis.}
\cref{tab:quality_threshold} examines the impact of quality score thresholds on pseudo-label selection. We observe a non-monotonic relationship: While lower thresholds ($\tau_{th} \leq 0.75$) generally yield superior performance by retaining more valid samples, excessively lenient selection ($\tau_{th}=0.50$) introduces noisy supervision, degrading results. The optimal balance occurs at $\tau_{th}=0.75$, achieving peak performance.

\begin{figure}[t]
    \centering
    \begin{subfigure}[b]{0.8\linewidth}
        \includegraphics[width=\linewidth]{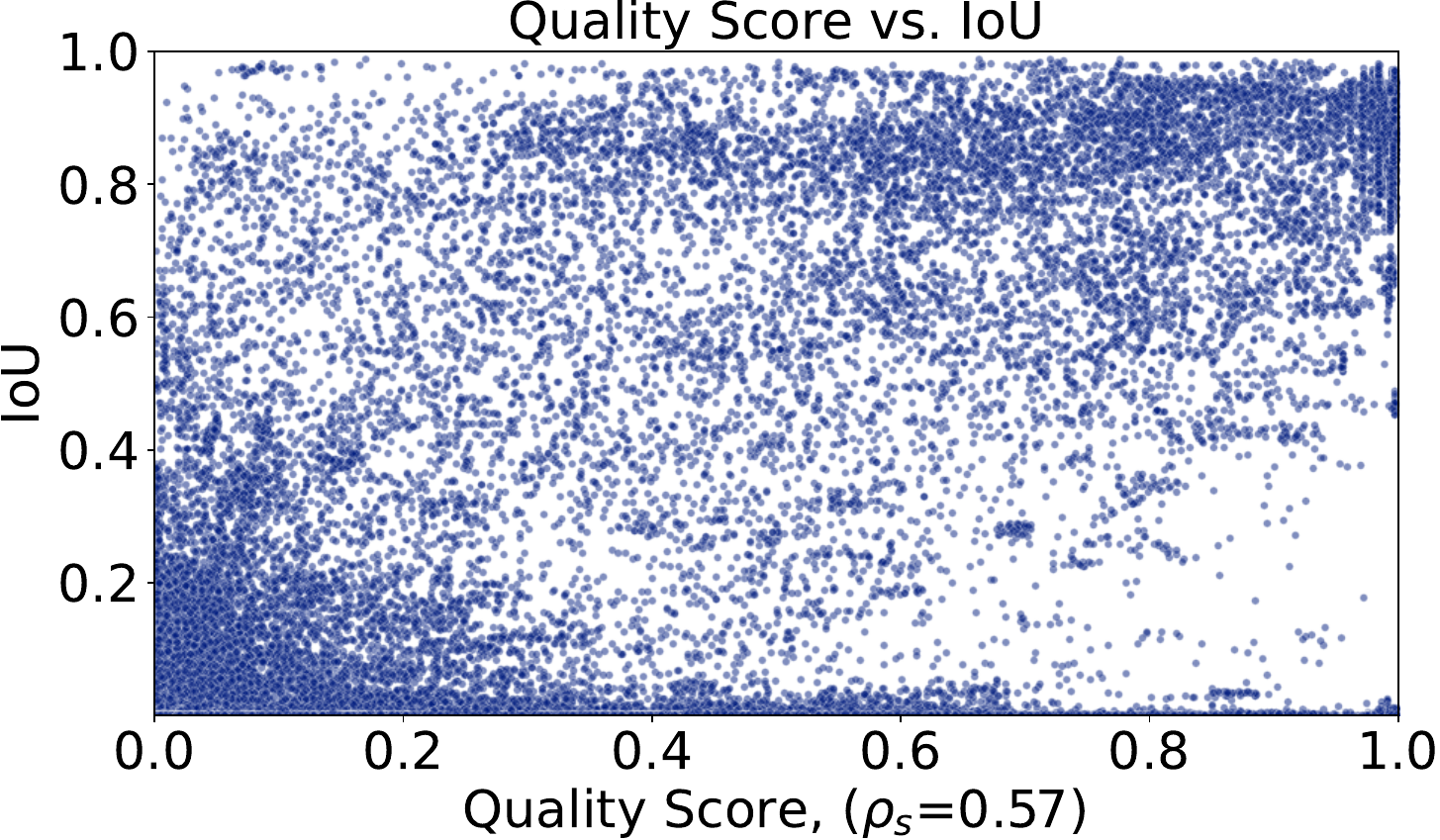}
        % \caption*{\parbox[c]{\linewidth}{\centering Quality score, $Q_l$ ($\rho_{s}=0.57$)}}
    \end{subfigure}
    \par\medskip
    \begin{subfigure}[b]{0.8\linewidth}
        \includegraphics[width=\linewidth]{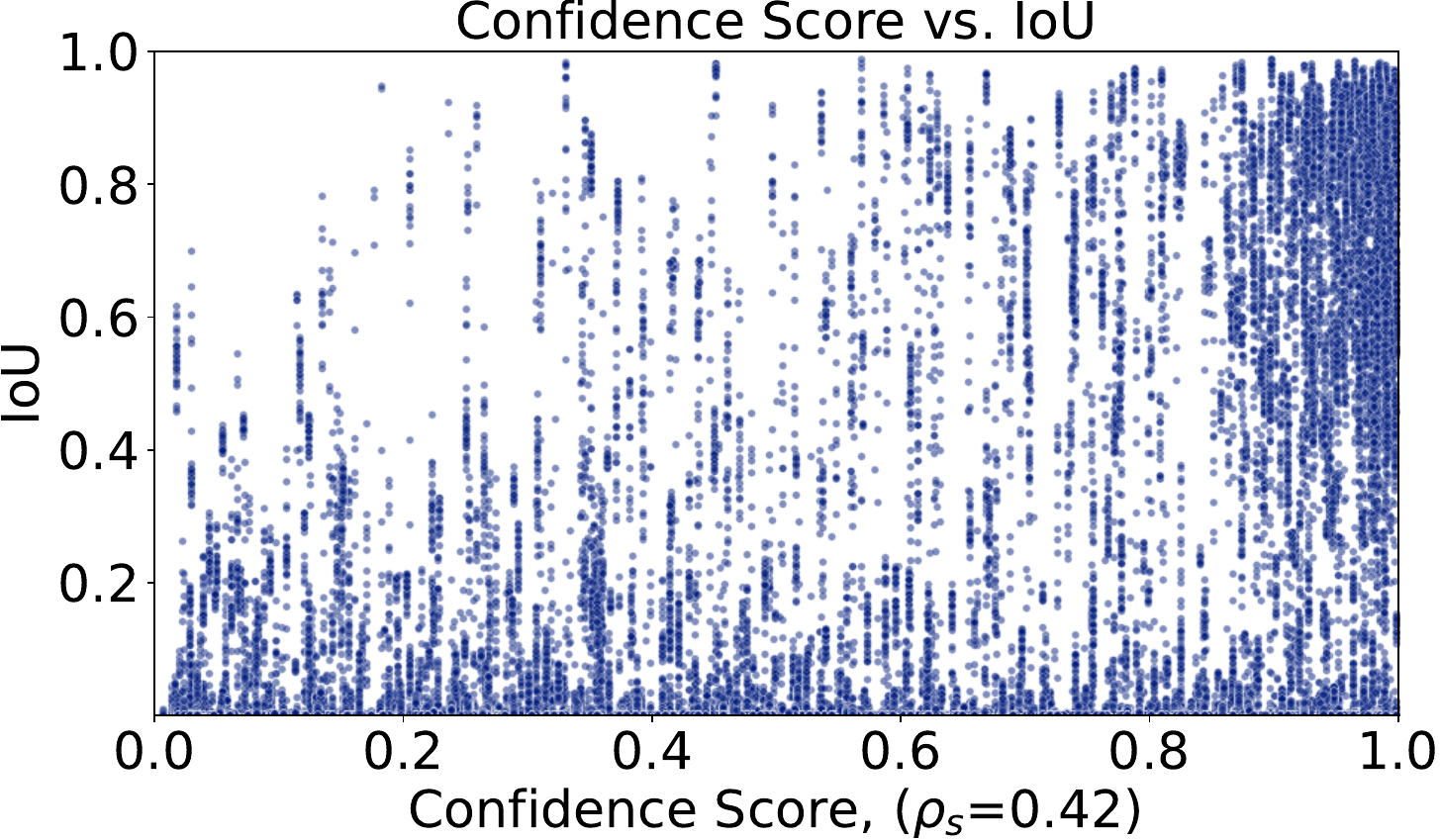}
        % \caption*{\parbox[c]{\linewidth}{\centering Confidence score, $s_l$ ($\rho_{s}=0.42$)}}
    \end{subfigure}
    % ONUR: 0.0 0.0 coincide, better origins
    \caption{
    \textbf{Visualized comparison of quality score $Q_l$ and confidence score $s_l$ on YouTubeVIS-2019 \texttt{val} split.} Here, $\rho_{s}$ denotes the Spearman's rank correlation coefficient. Subplot (a) visualizes quality scores $Q_l$ and their ground truth IoU. Subplot (b) visualizes confidence scores $s_l$ and their ground truth IoU. 
    %\todo{Make it bigger}
    }
    \label{fig:distribution}
\end{figure}

%\todo{I changed PDF to PNG for fast compilation. Don't forget to change this!}

\mypar{Quality score vs. confidence score.} As shown in \cref{fig:distribution},
%and \cref{tab:correlation}
our quality score $Q_s$ has a higher correlation to the IoU of the pseudo label and the ground truth label than the confidence score of VideoMask2Former~\cite{videomask2former}, which proves our quality predictor's effectiveness in pseudo-label quality assessment. This results in a significant improvement in the final VIS model performance (+2.1 $\text{AP}_{50}$) in \cref{tab:component_ablation}.

\subsection{Discussion}

% \begin{figure}[t]
%     \centering
%     \includegraphics[width=1\linewidth]{figures/area_vs_quality_score.pdf}
%     \caption{
%         \textbf{Correlation of the object area and the quality score on YouTubeVIS-2019 \texttt{val} split.} Here, \textit{Log Area} denotes the natural logarithm of the normalized area (object area divided by image area). $\rho_{s}$ denotes Spearman's rank correlation coefficient. As we can see, the quality predictor tends to give lower quality scores to smaller objects. \kaixuan{Where should I put the analysis of this figure?} \todo{change the log? add small, medium, big threshold line}
%     }
%     \label{fig:quality_score_and_area}
% \end{figure}

\begin{figure}[t]
    \centering
    \begin{subfigure}[b]{0.8\linewidth}
        \includegraphics[width=1\linewidth]{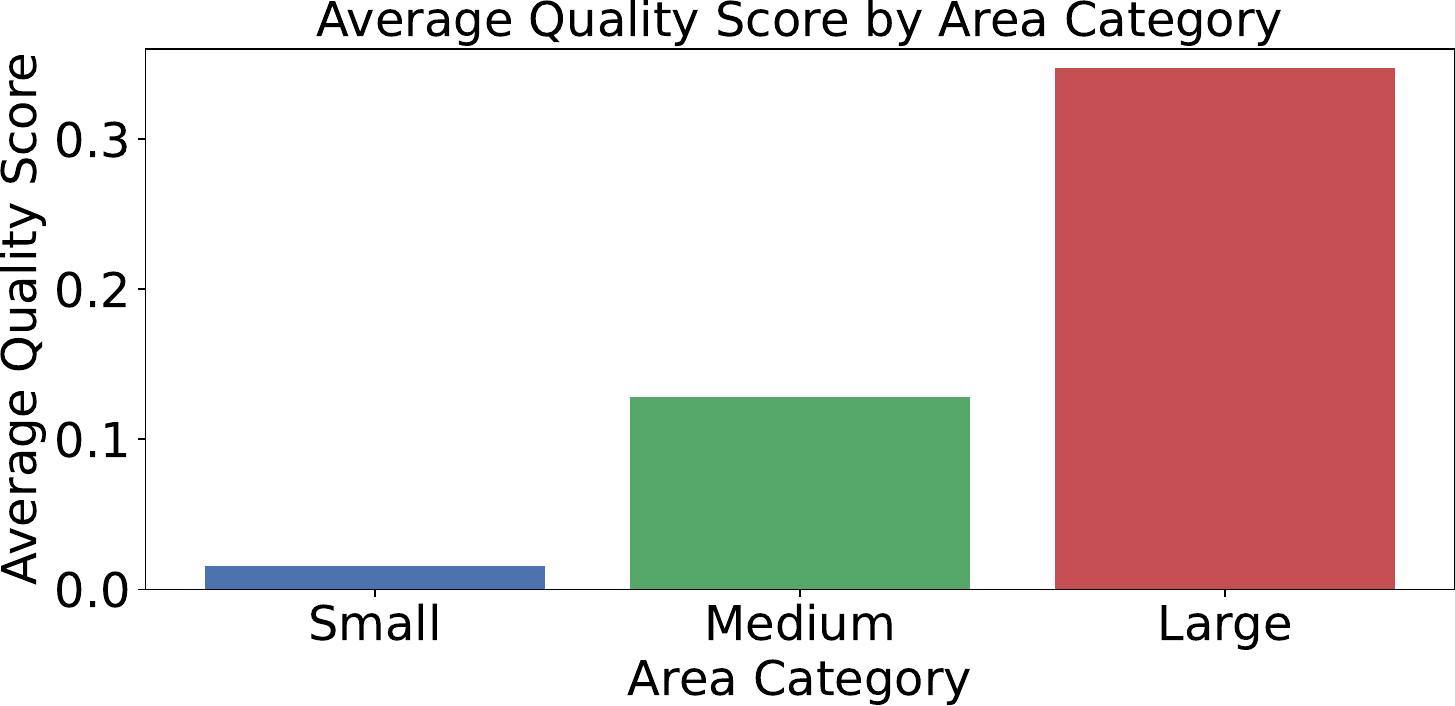}
    \end{subfigure}
    \par \medskip
    \begin{subfigure}[b]{0.8\linewidth}
        \includegraphics[width=1\linewidth]{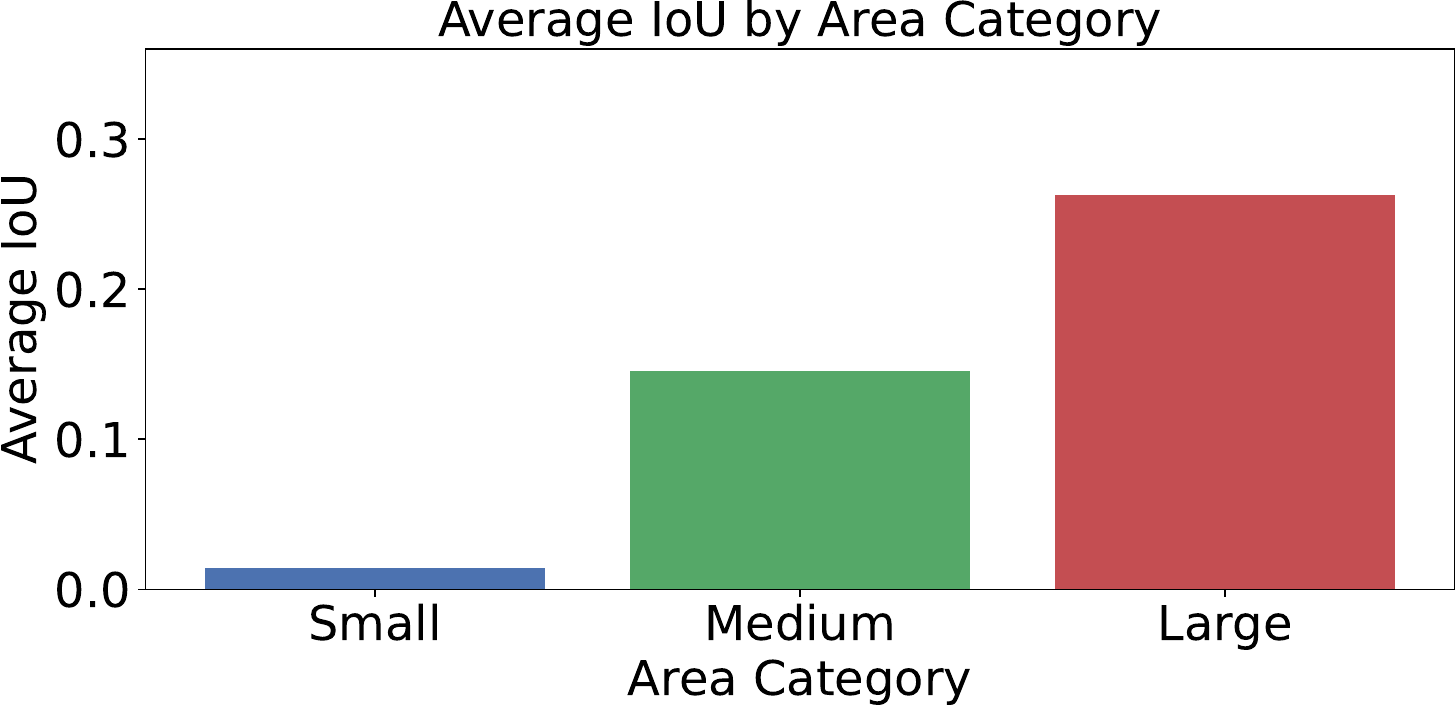}
    \end{subfigure}
    
    \caption{
        \textbf{Comparison of average ground-truth IoU and quality score across different mask sizes.} The definitions of \textit{Small}, \textit{Medium}, and \textit{Large} follow those in COCO~\cite{coco}. Here, \textit{IoU} refers to the intersection-over-union between the predicted mask and the ground-truth annotation. We compare the average IoU and quality score across these area categories. The results demonstrate a strong alignment between our quality score and the ground-truth IoU, with only a limited overestimation for large objects. This minor bias does not substantially affect the distribution of pseudo labels across training rounds (as shown in~\cref{fig:area_round_comparison}).
    }
    \label{fig:area_class_barplot}
\end{figure}

\begin{figure}[t]
    \centering
    \includegraphics[width=1.0\linewidth]{figures/area_distribution_all_rounds_kde.pdf}
    \caption{
        \textbf{Comparison of object area distributions between pseudo labels across different rounds and ground-truth labels.} Here, \textit{Log Area} refers to the natural logarithm of the normalized area (i.e., object area divided by image area). The definitions of \textit{Small}, \textit{Medium}, and \textit{Large} follow those in COCO~\cite{coco}. \textit{Round 0} denotes pseudo labels produced by the model in the beginning of the self-training. As shown, the object area distributions remain largely stable across different rounds. Nonetheless, relative to the ground truth, our pseudo-label set exhibits a higher proportion of large objects.
        %\kaixuan{Where should I put the analysis of this figure?} \todo{also show the gt}
    }
    \label{fig:area_round_comparison}
\end{figure}

% \paragraph{Alignment Between Quality Score, IoU, and Area Distributions.}
\mypar{Alignment Between Quality Score, IoU, and Area Distributions.}
As shown in~\cref{fig:area_class_barplot,fig:area_round_comparison}, our quality score exhibits a strong alignment with ground-truth IoU across different object sizes, reflecting mask reliability with only a limited overestimation for large instances. This consistency ensures that the confidence assigned to pseudo labels is well aligned with their true segmentation quality. Furthermore, the area distribution analysis across self-training rounds (\cref{fig:area_round_comparison}) indicates that the pseudo-label set remains stable, without substantial drift over iterations. Compared to ground-truth annotations, however, pseudo labels display a bias toward larger objects, which is consistent with the slight overestimation observed in quality scores. Importantly, this bias remains limited and does not hinder the effectiveness of iterative self-training.

\begin{table}[t]
\resizebox{\columnwidth}{!}{%
\begin{tabular}{lrrrrrrr}
\hline
\multicolumn{1}{c}{\multirow{2}{*}{\textbf{Method}}} & \multicolumn{7}{c}{\textbf{UVO-Dense}}      \\ \cline{2-8} 
\multicolumn{1}{c}{} & \textbf{${\text{AP}}_{50}$} & \textbf{${\text{AP}}_{75}$} & \textbf{${\text{AP}}$} & \textbf{${\text{AP}}_{S}$} & \textbf{${\text{AP}}_{M}$} & \textbf{${\text{AP}}_{L}$} & \textbf{${\text{AR}}_{10}$} \\ \hline
CutLER~\cite{cutler}                       & 10.3 & 3.4 & 4.6 & 1.3 & 11.5 & 17.3 & 8.8 \\
VideoCutLER~\cite{videocutler}                       & 13.5 & 5.1 & 6.2 & 1.7 & 14.8 & 26.5 & 12.3 \\
AutoQ-VIS                                            & 14.6 & 5.8 & 7.0 & 1.9 & 16.0  & 28.9 & 12.3  \\
\textit{vs. prev. SOTA}                              & {\color[HTML]{009901} +1.1}    & {\color[HTML]{009901} +0.7} & {\color[HTML]{009901} +0.8} & {\color[HTML]{009901} +0.2} & {\color[HTML]{009901} +1.2} & {\color[HTML]{009901} +2.4} & {\color[HTML]{009901} +0.0} \\ \hline
\end{tabular}%
}
\caption{
    \textbf{UVO-Dense \texttt{val}.} We reproduced the state-of-the-art VideoCutLER~\cite{videocutler} with the official code and checkpoint. AutoQ-VIS outperforms the previous state-of-the-art VideoCutLER by 1.1 $\text{AP}_{50}$. We evaluate results on UVO-Dense's \texttt{val} split in a class-agnostic manner.
}
\label{tab:uvo_dense}
\end{table}

% \paragraph{Generalizability.}
\mypar{Generalizability.}
To further assess the generalizability of our pipeline, we conduct experiments on UVO-Dense~\cite{uvo}. Following the same setup as in YouTubeVIS-2019, we train our model on synthetic videos from VideoCutLER~\cite{videocutler} together with the unlabeled \texttt{train} split of UVO-Dense, and evaluate it on the \texttt{val} split. UVO-Dense comprises 759 videos sourced from Kinetics-400~\cite{kay2017kinetics}, annotated at 30 FPS (503 for training and 256 for validation). Compared to YouTubeVIS-2019, UVO-Dense is considerably more challenging: it contains seven times more instance annotations per video and features crowded scenes with complex background motions. The training procedure remains identical to that used for YouTubeVIS-2019~\cite{ytvis2019}, except that we lower the quality score threshold to 0.3 to accommodate the higher object density in UVO-Dense.

% Table~\ref{tab:uvo_dense}
\Cref{tab:uvo_dense} compares AutoQ-VIS with prior state-of-the-art methods reported in~\cite{uvo}. AutoQ-VIS achieves a consistent improvement of about 1.1\% in $\text{AP}_{50}$, highlighting the robustness and generalizability of our pipeline.

\mypar{Evaluation Scope and Protocol Differences.}
We do not report results on YouTubeVIS-2021~\cite{ytvis2019}, OVIS~\cite{ovis}, or similar benchmarks because they do not release ground-truth annotations for the \texttt{val} split, and class-agnostic VIS evaluation requires access to these labels. We also clarify why our reproduced VideoCutLER~\cite{videocutler} baseline (48.2 ${\text{AP}}_{50}$) differs from the number in its original paper (50.7 ${\text{AP}}_{50}$): VideoCutLER reports results on the \texttt{train} split, whereas we evaluate on the \texttt{val} split, following same standard VIS evaluation practice.

%% file: sec/5_conclusion.tex
\section{Conclusion}

We present AutoQ-VIS, a quality-aware self-training framework that advances unsupervised video instance segmentation through iterative pseudo-label refinement with automatic quality control. By establishing a closed-loop system of pseudo-label generation and automatic quality assessment, our method achieves state-of-the-art performance (52.6 $\text{AP}_{50}$) on YouTubeVIS-2019 \texttt{val} split without requiring any human annotations. The simple quality predictor proves effective in pseudo-label quality assessment. 
% Future work may improve pseudo-label selection via temporal-aware quality scoring that leverages inter-frame consistency.    

% \onur{I did not like the way you formulate the main contribution: "bridging the synthetic-to-real domain gap". This sounds like a domain adaptation paper but our method is about sth. else. Can we rephrase it?}

% Current limitations stem from two aspects: (1) Hyperparameter sensitivity—the fixed quality threshold $\tau_{th}$ and the number of self-training rounds require careful tuning across datasets; (2) Simplistic quality modeling—our predictor currently evaluates frame-level mask quality without considering temporal consistency. Future work may explore temporal-aware quality scoring and online threshold adaptation to further improve performance.

% \onur{refine the references, inconsistent formating}

%% file: sec/X_suppl.tex
\clearpage
\setcounter{page}{1}
\maketitlesupplementary
%\todo{More results, maybe video, two pages like videocutler, what is selected high medium low quality, visual examples. show more examples per line 90\% 70\% 50\% ... so people can see what quality selector does}

% \section{Rationale}
% \label{sec:rationale}
% % 
% Having the supplementary compiled together with the main paper means that:
% % 
% \begin{itemize}
% \item The supplementary can back-reference sections of the main paper, for example, we can refer to \cref{sec:intro};
% \item The main paper can forward reference sub-sections within the supplementary explicitly (e.g. referring to a particular experiment); 
% \item When submitted to arXiv, the supplementary will already included at the end of the paper.
% \end{itemize}
% % 
% To split the supplementary pages from the main paper, you can use \href{https://support.apple.com/en-ca/guide/preview/prvw11793/mac#:~:text=Delete%20a%20page%20from%20a,or%20choose%20Edit%20%3E%20Delete).}{Preview (on macOS)}, \href{https://www.adobe.com/acrobat/how-to/delete-pages-from-pdf.html#:~:text=Choose%20%E2%80%9CTools%E2%80%9D%20%3E%20%E2%80%9COrganize,or%20pages%20from%20the%20file.}{Adobe Acrobat} (on all OSs), as well as \href{https://superuser.com/questions/517986/is-it-possible-to-delete-some-pages-of-a-pdf-document}{command line tools}.

\section{Detailed methodology}
\label{detailed_method}

This section supplements what is not clearly stated in \cref{sec:multi_round_training}.

\subsection{Automated pseudo-annotation with spatio-temporal NMS}

After the training of the VIS model, we use it to label the unlabeled videos. 
Let $\mathcal{D} = \{d_i\}^{N}_{i=1}$ denote the initial detection set per video, where each detection $d = (s_i, \{m_i^t\}^{T}_{t=1})$ contains:
\begin{itemize}
    \item $s_i \in [0,1]$: Confidence score
    \item $\{m_i^t\}^{T}_{t=1}$: Binary mask sequence across $T$ frames
\end{itemize}

We filter those detections that have confidence scores larger than or equal to 0.25:
\begin{equation}
    \mathcal{D}_{\text{filtered}} = \{ d_i \in \mathcal{D} \mid s_i \geq 0.25 \}
\end{equation}

However, the detection sets may contain duplicate detections. To solve this problem, we need to perform spatiotemporal non-maximum suppression. First, we sort the detections based on their confidence scores:
\begin{equation}
    \mathcal{D}_{\text{sorted}} = \{ d_{(k)} \}_{k=1}^{|\mathcal{D}_{\text{filtered}}|} \quad \text{s.t.} \quad \forall i < j: s_{(i)} \geq s_{(j)}
\end{equation}

Our method eliminates redundant detections through spatiotemporal overlap analysis: Any lower-confidence prediction is suppressed if exhibiting mask overlap (IoU $\geq$ 0.5) with higher-confidence detections in at least one video frame. Formally, let $\mathcal{D}_{\text{suppressed}} \subseteq \mathcal{D}_{\text{sorted}}$ represent the preserved detection set after suppression:

\begin{equation}
\begin{split}
d_{(k)} \in \mathcal{D}_{\text{suppressed}} \iff & \nexists d_{(p)} \in \mathcal{D}_{\text{suppressed}} \ \text{where} \ p < k \ \text{s.t.} \\
& \exists t \in [1,T] : \frac{\|m_{(k)}^t \cap m_{(p)}^t\|}{\|m_{(k)}^t \cup m_{(p)}^t\|} \geq 0.5 \\
& \underbrace{\hphantom{\exists t \in [1,T] : \frac{\|m_{(k)}^t \cap m_{(p)}^t\|}{\|m_{(k)}^t \cup m_{(p)}^t\|}}}_{\mathclap{\text{Frame-specific overlap condition}}}
\end{split}
\end{equation}
where $\| \cdot \|$ denotes pixel cardinality. This temporal-existential criterion suppresses duplicates appearing in any frame of the video sequence.

\subsection{Confidence-aware filtration via quality predictor}

Let $\mathcal{D}_{\text{global}}^{(k)}$ denote the union of $\mathcal{D}_{\text{suppressed}}$ of all videos:
\begin{equation}
    \mathcal{D}_{\text{global}}^{(k)} = \bigcup_{v \in \mathcal{V}} \mathcal{D}_{\text{suppressed},v}^{(k)}
\end{equation}
where $\mathcal{D}_{\text{suppressed},v}^{(k)}$ denotes preserved detections for video $v$ after spatiotemporal NMS at iteration $k$.

For each detection $d \in \mathcal{D}_{\text{global}}^{(k)}$, we define:
\begin{equation}
    Q_d^t = s_d \cdot \hat{\text{IoU}}_d^t \quad \forall t \in [1,T_v]
\end{equation}
where $Q_d^t$ is the quality score of detection $d$ in frame $t$, $s_d$ is the confidence score of detection $d$, and $T_v$ denotes the number of frames in video $v$.

We implement quality-based pseudo-label selection using a fixed quality threshold $\tau_{\text{th}}$. For each detection $d \in \mathcal{D}_{\text{global}}^{(k)}$ across all videos:

\begin{equation}
    \mathcal{S}_d^t = \begin{cases} 
1 & Q_d^t \geq \tau^{(k)} \\
0 & \text{otherwise}
\end{cases}
\end{equation}
where $\mathcal{S}_d^t$ denote whether pseudo-label of detection $d$ in frame $t$ is selected. For each detection, if its results are not selected in any frames, we discard it:

\begin{equation}
\mathcal{D}_{\text{retained}}^{(k)} = \left\{ d_v \in \mathcal{D}_{\text{global}}^{(k)} \,\bigg|\, \sum_{t=1}^{T_v} \mathcal{S}_d^t > 0 \right\}
\end{equation}
where $\mathcal{D}_{\text{retained}}^{(k)}$ is the set of detections we retain in iteration $k$.

\section{Additional qualitative visualizations}

\begin{figure*}[h]
    \centering
    \includegraphics[width=1\linewidth]{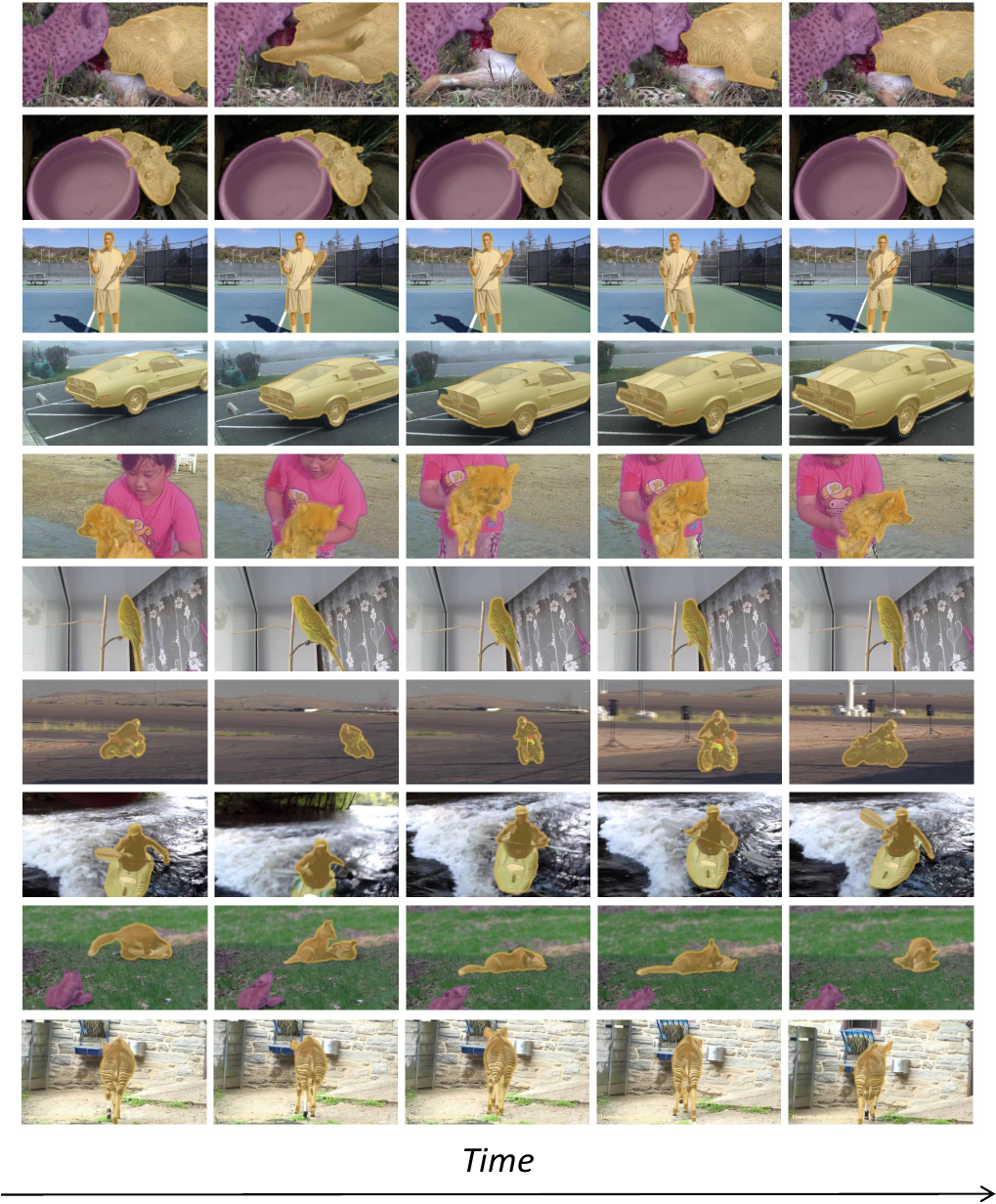}
    \caption{\textbf{Qualitative results of our VIS model on YouTubeVIS-2019 \texttt{val} split.}}
    \label{fig:add_vis_visual}
\end{figure*}

\begin{figure*}[h]
    \centering
    \includegraphics[width=1\linewidth]{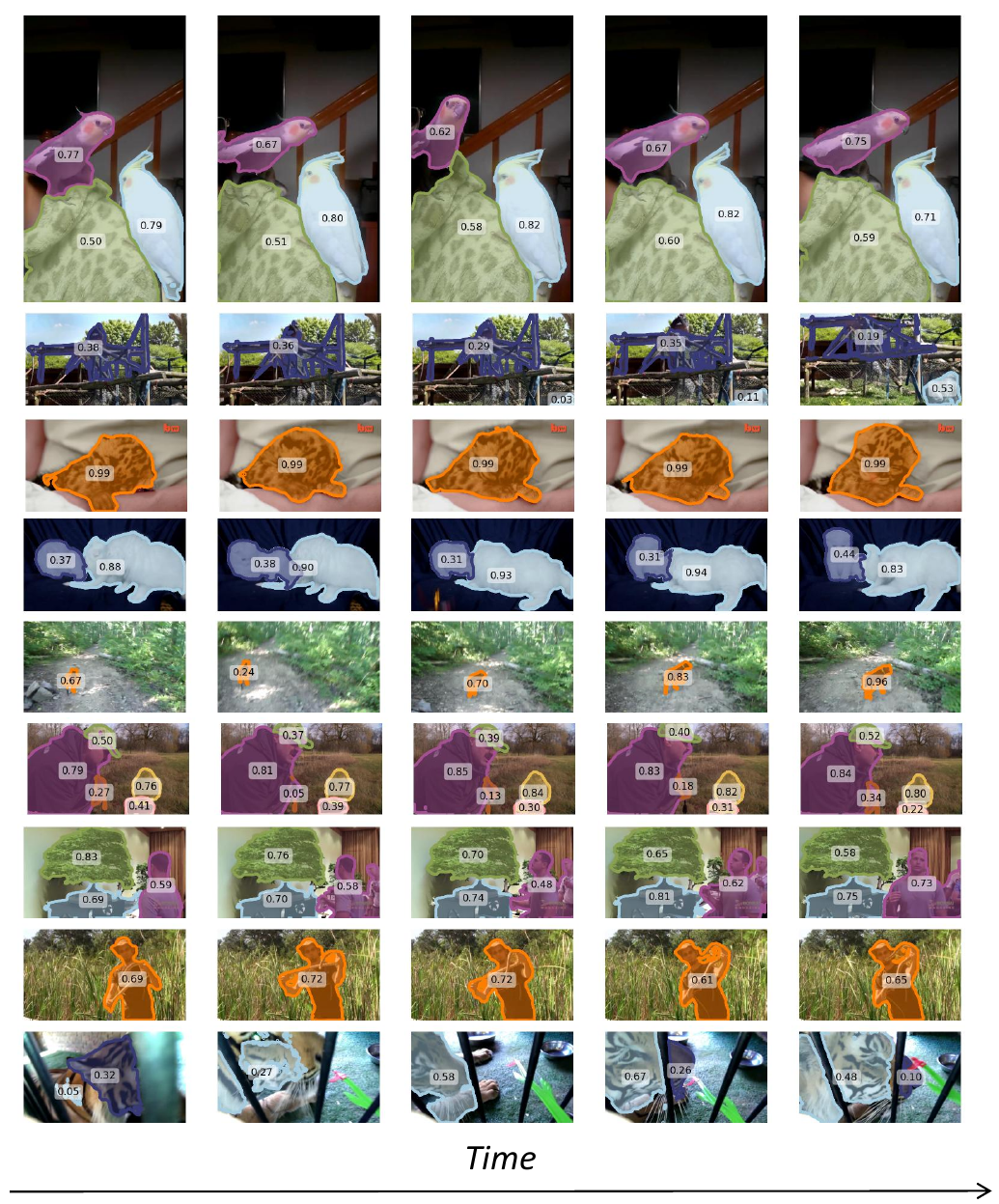}
    \caption{\textbf{Qualitative results of our quality predictor on YouTubeVIS-2019 \texttt{train} split.} The quality scores are shown in the center of each pseudo label.}
    \label{fig:quality_visual}
\end{figure*}

We provide additional qualitative results of our VIS model and quality predictor in \cref{fig:add_vis_visual} and \cref{fig:quality_visual}.